\pdfoutput=1

\documentclass[lettersize,journal]{IEEEtran}

\usepackage{amsfonts}
\usepackage{amsmath}
\usepackage{multirow}
\usepackage{booktabs}
\usepackage{siunitx}
\usepackage{graphicx}
\usepackage{gensymb}
\usepackage{tablefootnote}

\usepackage{tabularx}
\usepackage{makecell}
\usepackage{url}
\usepackage[noadjust]{cite}

\usepackage{comment}
\usepackage{amssymb}
\usepackage{algorithm}
\usepackage{algpseudocode}
\usepackage{pifont}

\usepackage{threeparttable}
\usepackage[caption=false]{subfig}

\newcommand{\best}[1]{\textbf{#1}}

\title{\LARGE \bf{FGSCalib}: Targetless Camera--LiDAR Extrinsic Calibration from Foreground Segmentation}

\author{Marcell K\'egl, Csaba Benedek
\thanks{
This work was supported in part by the TKP2021-NVA-01 and OTKA \#143274 projects of the Hungarian NRDI Office.}
 \thanks{M. K\'egl and C. Benedek are with the Hungarian Research Network Institute for Computer Science and Control (HUN-REN SZTAKI), 1111 Budapest, Kende utca 13-17, and with the P\'azm\'any P\'eter Catholic University, Faculty of Information Technology and Bionics, 1083 Budapest, Pr\'ater utca 50/A. (e-mail: kegl.marcell@sztaki.hun-ren.hu, benedek.csaba@sztaki.hun-ren.hu).
}
}

\usepackage{eso-pic}
\AddToShipoutPictureBG*{\AtPageUpperLeft{\raisebox{-7mm}{\makebox[\paperwidth]{\footnotesize This work has been submitted to the IEEE for possible publication. Copyright may be transferred without notice, after which this version may no longer be accessible.}}}}
\begin{document}

\maketitle
\bstctlcite{BSTcontrol}

\begin{abstract}
Camera--LiDAR pairs installed as fixed infrastructure need periodic recalibration, but a calibration target takes the installation out of service. We present FGSCalib, a targetless method that calibrates a stationary camera--LiDAR pair from the objects moving in its view. We only assume that arbitrary moving objects are present in the scene: each sensor separates the moving foreground from the static background, without relying on object classes. With the correct extrinsic, the LiDAR foreground points project onto the camera foreground. FGSCalib turns the camera mask into a boundary-proximity field and finds the extrinsic by two-stage simulated annealing over selected frame pairs. It needs no measured initial extrinsic (random starts give similar results) and works with both rotating and non-repetitive LiDARs. We evaluate it on 21 simulated scenarios with exact ground truth and on 72 real ones (nine camera--LiDAR pairs in eight scenes), with a reference averaged from 43 manual calibrations. In simulation the median error is 5.3\,cm and $0.19^{\circ}$. On real data we use a tolerance of 18\,cm and $1.8^{\circ}$, met by 93\% of single manual calibrations. Keeping the lowest-cost of five runs, FGSCalib meets it in 51 of the 72 scenarios: in 20 of 24 with the Livox, where the strongest of five baselines reaches 11, and in 31 of 48 with the two rotating LiDARs, where only one baseline estimate does. Its median error is 9.5\,cm and $0.9^{\circ}$, against 7.3\,cm and $0.7^{\circ}$ for a single manual calibration. The code and the dataset are available online.
\end{abstract}


\begin{figure*}[t]
\centering
\includegraphics[width=0.94\textwidth]{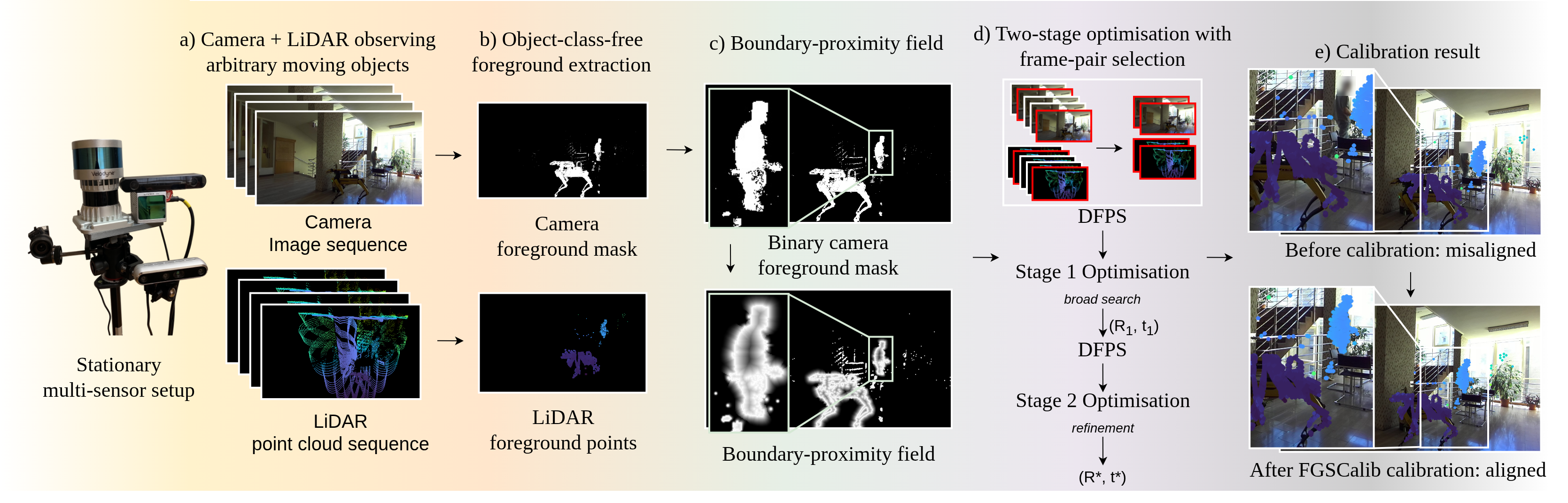}
\caption{FGSCalib pipeline. (a) A stationary camera--LiDAR rig records moving objects. (b) Foreground is segmented in both sensors without object classes: a binary camera mask and a set of LiDAR foreground points. (c) The camera mask is converted into the boundary-proximity field. (d) Frame pairs are chosen by diversity-aware frame-pair selection (DFPS) and the extrinsic is searched by simulated annealing in two stages, a broad search and a refinement. (e) LiDAR foreground points projected into the image before and after calibration.}
\label{fig:pipeline_overview}
\end{figure*}

\section{Introduction}
\label{sec:introduction}

Camera--LiDAR pairs are increasingly mounted at fixed vantage points to observe moving scenes, for traffic and infrastructure monitoring~\cite{zhang2026roadside_review}, public-space sensing~\cite{snapcat3d} and surveillance~\cite{ISPRS}. Integrating the visual information captured by the camera with the geometric data of the LiDAR (light detection and ranging) sensor requires the rigid extrinsic transformation between their coordinate systems, whose error every fused measurement inherits~\cite{calibration_review}. Vibration, thermal cycling and occasional disturbance change this transformation over the life of an installation~\cite{tian2024rlcformer}, and re-establishing it with a calibration target means bringing the target to the mount and taking the installation out of service. What such sites do have, continuously, is moving objects in view; this paper turns them into the calibration signal.

Existing methods are less well suited to this setting. Target-based methods match the known geometry of artificial markers across the sensors~\cite{OLIVEIRA2022ATOM,Pusztai_boxcalib_2017,tekla_calibsphere_2020}; they are accurate but need manual effort and controlled conditions~\cite{semalign}. Targetless methods exploit scene features~\cite{koide2023general,hku_calib,liu2022hku_adaptive_voxelization}, semantic segmentation~\cite{semalign,sst-calib,autos_n1}, motion trajectories~\cite{nagyB_2020_calib,motion_based_calib,sfm_balazs_nagy,nagy2019online_e2e_calib} or learned representations~\cite{regnet,lccnet}, and most of them rely on static scene content through which the sensor moves~\cite{zhang2025lcc_review}. A stationary installation never moves, and its static background may have little structure, such as a plain wall, or be hidden by the moving objects that dominate its view. Few methods use moving objects seen by stationary sensors: Jang et al.~\cite{jang2025trafficcalib} calibrate a traffic-surveillance camera--LiDAR pair from the edges of passing vehicles, and Glas et al.~\cite{snapcat3d} a network of range sensors from tracked pedestrians.

In this paper, we introduce FGSCalib (Foreground-Segmentation Calibration), a method for estimating the extrinsic calibration of a stationary camera--LiDAR pair using moving objects observed simultaneously by both sensors. Moving objects appear as foreground pixels in the camera images and as foreground points in the corresponding LiDAR sweeps. When the sensors are correctly calibrated, the projected LiDAR foreground points should spatially align with the foreground regions in the camera images.

Unlike approaches based on object tracking or trajectory correspondence, FGSCalib does not require explicit object associations. Instead, we employ background modelling to extract foreground information independently from both sensor modalities, without relying on predefined object classes. For LiDAR foreground extraction, the method adopts the model introduced in our previous work~\cite{ISPRS}. The extrinsic transformation is then estimated by optimising the alignment between the camera and LiDAR foreground observations across multiple time instances (Fig.~\ref{fig:pipeline_overview}).

LiDAR-based foreground detection faces challenges that differ between LiDAR types. Rotating multi-beam (RMB) LiDARs acquire measurements at fixed elevation angles, producing sparse and repetitive scan rings when the sensor remains stationary. In contrast, non-repetitive circular scanning (NRCS) LiDARs follow a scanning pattern that progressively increases the spatial coverage over time, enabling the accumulation of high-resolution point clouds without moving the sensor. However, achieving sufficient temporal resolution for detecting moving objects requires short integration times, resulting in sparse foreground measurements. Thus, foreground extraction involves balancing spatial coverage and temporal resolution while maintaining an accurate background model, as introduced in~\cite{ISPRS}.

The methods presented in~\cite{koide2023general,hku_calib} require dense point clouds, which, for RMB LiDARs, can only be obtained by moving the sensor during data acquisition. However, sensor motion prevents the reliable detection of moving objects through background modelling. Therefore, FGSCalib performs foreground--background segmentation directly on the sparse LiDAR point clouds acquired by a stationary sensor, in which moving objects are still sampled as they cross the scan rings.

We evaluate the proposed approach using three LiDAR sensors representing two distinct scanning principles: the Velodyne VLP-16 and the Ouster OS1-64, which are RMB LiDARs with 16 and 64 scan rings, respectively, and the Livox Avia, an NRCS LiDAR.

FGSCalib assumes that (i) the camera and the LiDAR are rigidly mounted with fields of view that overlap where objects move; (ii) the camera intrinsics are known; (iii) each camera image can be paired with the LiDAR sweep nearest to it in time, giving a sequence of \emph{frame pairs}; and (iv) the recording contains moving objects. The search needs no measured initial extrinsic, only a box of poses that contains the true one. We start it from the \emph{nominal pose} (zero translation, both sensors facing the same way); random starts in the box give similar results (Sec.~\ref{sec:manual}). Real recordings have no ground truth, so we build a reference from repeated manual calibrations and derive the tolerance from how well they agree (Secs.~\ref{sec:reference} and~\ref{sec:evaluation}).

The main contributions of this work are:
\begin{itemize}
    \item FGSCalib, a targetless calibration method for stationary camera--LiDAR pairs that uses moving foreground as its signal, through a boundary-proximity field derived from the camera mask and a two-stage simulated-annealing search over a box of poses. It needs no target, platform motion, object class or measured initial extrinsic (random starts anywhere in the box give results similar to the nominal start), and it works with both RMB and NRCS LiDARs.
    \item Diversity-aware frame-pair selection (DFPS), a target-free rule that chooses the frame pairs used by the search from the foreground segmentations of the two sensors alone; in our ablation it lowers the largest rotation errors.
    \item A new benchmark of 21 simulated and 72 real scenarios with three LiDARs of two scanning principles (RMB and NRCS) and three cameras, with 43 manual calibration sessions that give the real-data reference, its uncertainty, and a tolerance derived from how closely single sessions agree. The benchmark is released with this paper. To our knowledge, it is the first public camera--LiDAR benchmark that provides several independent manual calibrations per sensor pair, which are needed to measure the uncertainty of the reference.
    \item An evaluation against manual calibration and five targetless baselines, with a multi-seed ablation: on real data, with the lowest-cost of five runs, FGSCalib is within the tolerance in 20 of 24 scenarios with the Livox and in 31 of 48 with the RMB LiDARs, where the baselines reach 11 and 1; on static recordings, where FGSCalib cannot run, a static-scene baseline reaches it in 11 of 12.
\end{itemize}

\section{Related Work}
\label{sec:related_work}

Surveys of camera--LiDAR calibration are given in~\cite{calibration_review,zhang2025lcc_review}; here we focus on targetless methods.

\subsection{Methods That Do Not Use Scene Motion}

Most targetless methods rely on static scene structure and assume that the scene remains unchanged across the calibration frames, with any apparent motion caused only by the sensor platform. Moving objects therefore contribute no useful information to them and can disrupt the correspondences they rely on.

\paragraph{Feature-based methods} These methods extract geometrically corresponding primitives from both modalities. Koide et al.~\cite{koide2023general} convert LiDAR intensity measurements into a virtual image and use SuperGlue~\cite{sarlin2020superglue} to establish 2D--3D keypoint correspondences, which are then refined by normalised information distance. The method requires LiDAR-intensity texture and a dense point cloud, so an RMB LiDAR must be moved during the recording to accumulate enough points. Yuan et al.~\cite{hku_calib} match depth-continuous point-cloud edges against Canny image edges under an explicit LiDAR noise model; the method reaches pixel-level accuracy in scenes with well-defined planar edges but requires an initial estimate of the extrinsic.

\paragraph{Segmentation-based methods} These methods align corresponding regions rather than sparse features. Zhu et al.~\cite{autos_regi} segment obstacle classes in the camera image and turn the mask into a smooth map, which they call a height map, decaying inside and outside each obstacle region under an exponential kernel of the Manhattan ($\ell_1$) distance; an extrinsic is scored by the map values at the projected non-ground LiDAR points. This formulation is the most closely related to our proposed boundary-proximity field, with the key differences discussed in Sec.~\ref{sec:dt_invdt}. Sun et al.~\cite{autos_n1} associate vehicle instances across the two modalities with the Hungarian algorithm and score an extrinsic by a distance-transform mask match plus instance-centroid distances, refining an initial estimate from hand--eye calibration. SemAlign~\cite{semalign} and SST-Calib~\cite{sst-calib} optimise differentiable losses over same-class projections. These four methods all require predefined semantic classes to be present in the scene.

Calib-Anything~\cite{calibanything} and PAIRS-Calib~\cite{pairs2026} eliminate the need for predefined object classes by employing class-agnostic segmentation. Calib-Anything uses the Segment Anything Model to segment the camera images and refines an initial calibration estimate by maximising the consistency of the projected LiDAR points within each segment in terms of intensity, surface normals, and segment categories. The method is demonstrated primarily on static driving scenes. PAIRS-Calib aligns class-agnostic boundary masks extracted from camera images and from LiDAR depth and intensity representations using stochastic approximation, searching within a bounded region around an initial calibration estimate. It is evaluated on recordings acquired from a moving vehicle, with initial calibration errors of at most $2.5^\circ$ in rotation and 0.25\,m in translation. A direct experimental comparison with PAIRS-Calib was not possible, as its implementation has not been made publicly available.

\paragraph{Learning-based methods} Approaches such as CalibNet~\cite{calibnet}, RegNet~\cite{regnet}, LCCNet~\cite{lccnet}, and CMRNext~\cite{cmrnext}, which builds on RAFT~\cite{raft}, employ neural networks to estimate the extrinsic transformation directly or to establish dense pixel-to-point correspondences. Their publicly available models are primarily trained on driving datasets acquired with RMB LiDAR sensors. Although CMRNext has demonstrated generalisation to previously unseen sensors, the performance of learning-based approaches remains dependent on the distribution of their training data~\cite{huang2025whatmatters}. Adapting these methods to new sensor configurations typically requires retraining with substantial amounts of data annotated with ground-truth extrinsic parameters. In contrast, 3DGS-Calib~\cite{3dgs_calib} avoids the need for labelled training data by leveraging Gaussian Splatting, but requires camera motion within a static environment to construct its scene representation.

\paragraph{Ego-motion-based methods} These approaches estimate the extrinsic transformation by aligning the sensor ego-motion trajectories independently estimated from camera and LiDAR measurements as the sensor platform moves~\cite{nagyB_2020_calib,motion_based_calib,sfm_balazs_nagy,nagy2019online_e2e_calib}. However, moving objects in the environment may compromise the accuracy of these motion estimates, and the methods are inherently unsuitable for stationary sensor configurations.

\subsection{Methods Exploiting the Motion of Scene Objects}

A smaller subset of methods instead exploits the motion of objects in the scene to estimate the extrinsic calibration. Glas et al.~\cite{snapcat3d} calibrate a network of stationary 3D range sensors rather than a camera--LiDAR pair; they detect and track pedestrians across the sensors and align the resulting trajectories, which makes the method dependent on the presence of pedestrians in the scene. Peršić et al.~\cite{pervsic_trajectory_based} also exploit moving objects for calibration by tracking their motion from a mobile sensor platform. However, their approach requires explicit object tracking and estimates only the rotational component of the extrinsic transformation. Jang et al.~\cite{jang2025trafficcalib} calibrate a fixed traffic-surveillance camera--LiDAR pair from passing vehicles. Similarly to our approach, their method identifies moving objects in both sensor modalities through class-free background subtraction. It then extracts the edges of the vehicles and estimates the extrinsic transformation by maximising the kernel correlation between projected LiDAR edge points and camera edge pixels using a genetic algorithm. The optimisation is initialised with candidate transformations sampled within twice the initial calibration error, and the method is evaluated with initial errors of up to $10^\circ$ and 1\,m using 40- and 64-channel RMB LiDARs. Although both approaches exploit foreground information from moving objects, they differ in two important aspects. First, FGSCalib evaluates the alignment of entire foreground regions rather than relying exclusively on object boundaries, making use of foreground measurements even when sparse LiDAR sweeps provide limited boundary information. Second, FGSCalib optimises over a predefined pose range without requiring a measured initial calibration estimate, whereas~\cite{jang2025trafficcalib} defines its search space relative to an initial pose. These differences are particularly relevant with sparse LiDAR observations and limited prior calibration information. A direct experimental comparison was not possible because the implementation of~\cite{jang2025trafficcalib} is not publicly available.

While these approaches typically rely on additional assumptions or prior information, FGSCalib estimates the full camera--LiDAR extrinsic transformation from class-free foreground observations of moving objects, without requiring sensor motion, explicit object correspondences, or a measured initial extrinsic estimate.

\subsection{Data Selection for Calibration}

Camera--LiDAR calibration typically relies on multiple frame pairs collected over time. However, not all observations are equally informative for estimating the extrinsic transformation, and many existing approaches either use all available frames or select a subset randomly. In target-based calibration, Tsai et al.~\cite{Tsai2021OptimisingTS} address this issue by ranking chessboard observations according to their orientation diversity and detection accuracy. Their approach, however, relies on a known calibration target. In contrast, FGSCalib introduces a target-free frame selection strategy that scores each frame pair directly from the foreground segmentations of both sensors, favouring pairs that differ in where and at what depth the foreground appears (Sec.~\ref{sec:data_selection}).
\section{Methodology}
\label{sec:methodology}

FGSCalib has four steps (Fig.~\ref{fig:pipeline_overview}b--d): foreground segmentation in both sensors (Sec.~\ref{sec:Featureextraction}), the boundary-proximity field and the cost defined on it (Sec.~\ref{sec:dt_invdt}), frame-pair selection (Sec.~\ref{sec:data_selection}) and a two-stage simulated-annealing search (Sec.~\ref{sec:optimization}).

\subsection{Foreground Segmentation for Camera and LiDAR Data}
\label{sec:Featureextraction}

\paragraph{Camera} We use an improved Gaussian-mixture background model (mixture of Gaussians, MoG)~\cite{zivkovic2004improved_MOG2} with its standard settings, which classifies each pixel over time as foreground or background and gives a binary mask $M$ per frame. Pixels that MoG labels as shadow are treated as background. We apply no further post-processing to the masks, not even the removal of small connected foreground regions (\emph{components}), because the field of Sec.~\ref{sec:dt_invdt} limits the influence of small spurious components. The rest of the pipeline uses only these masks, so another foreground segmenter could in principle replace MoG.

\paragraph{LiDAR} We use the LiDAR foreground model of our previous work~\cite{ISPRS}, an adaptation of the Stauffer--Grimson mixture model~\cite{stauffer_grimson} to range measurements. An RMB LiDAR returns one range per elevation channel and azimuth step, so its output already forms a regular lattice in which each cell is one fixed viewing direction; the points of an NRCS LiDAR, or of an RMB LiDAR whose driver outputs an unordered cloud, are binned by azimuth and elevation into a lattice of the same form. Each cell keeps a mixture of five Gaussian modes over the ranges measured in its direction, updated online with each sweep, and the mode with the largest weight gives the background range of the cell. A return that matches this component is background and updates the model. Three rules adapt the model to moving objects and to sparse, jittering returns. A return clearly in front of the background range is foreground and leaves the model unchanged, so an object that stops is not absorbed into the background. A return that matches the background of any cell in its $3{\times}3$ neighbourhood is background, which absorbs beam jitter and, for the NRCS LiDAR, the binning. On the real data, a return close to a stable second mode of its cell is also background, as on object edges where a cell alternates between two surfaces. Returns closer than 1.2\,m are discarded, and the constants are in the supplement (Sec.~S2). The output is, for each frame pair $s$, a set of foreground points $P^L_s$ (Fig.~\ref{fig:pipeline_overview}b, Fig.~\ref{fig:real_field_b}), and a frame pair enters the \emph{candidate pool}, from which Sec.~\ref{sec:data_selection} selects, whenever its LiDAR foreground is non-empty.

\subsection{The Boundary-Proximity Field}
\label{sec:dt_invdt}
A LiDAR point $p$ has camera coordinates $Rp+t$, with the extrinsic rotation $R\in\mathrm{SO}(3)$ and translation $t\in\mathbb{R}^{3}$, and projects to the pixel $\operatorname{proj}(p,R,t)$ by the pinhole model with lens distortion, using the known intrinsics. The goal of calibration is to find the extrinsic $(R, t)$ for which the projected LiDAR foreground points fall on the camera foreground.

\begin{figure}[t]
\centering
\subfloat[Binary mask]{\includegraphics[width=0.48\linewidth]{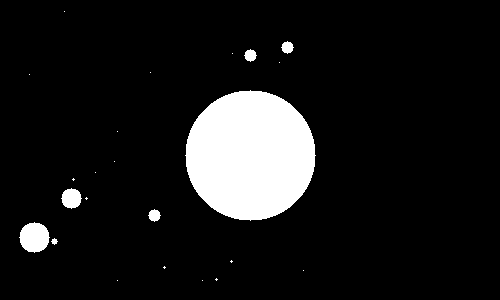}\label{fig:dt_mask}}%
\hfill
\subfloat[The field (ours)]{\includegraphics[width=0.48\linewidth]{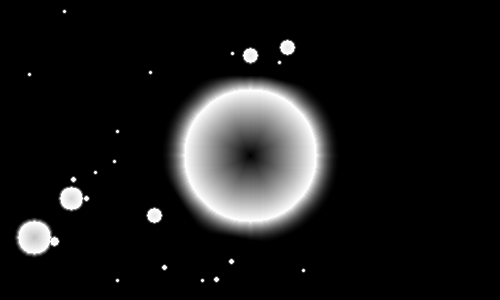}\label{fig:dt_merged_ccomp}}\\[2pt]
\subfloat[The field on a real recording]{\includegraphics[width=0.48\linewidth]{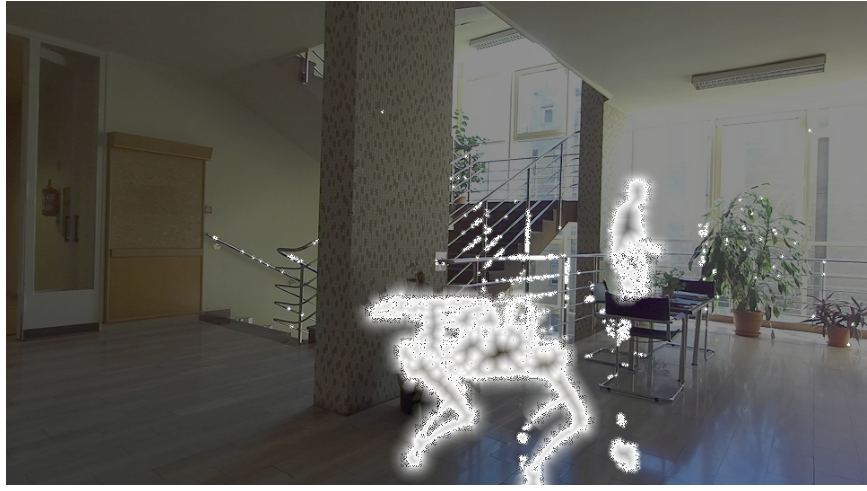}\label{fig:real_field_a}}%
\hfill
\subfloat[The same frame with the LiDAR foreground]{\includegraphics[width=0.48\linewidth]{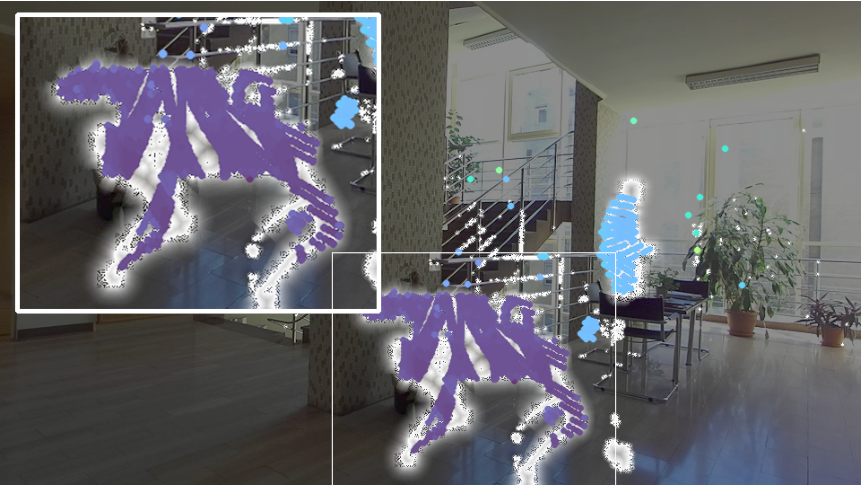}\label{fig:real_field_b}}
\caption{The boundary-proximity field (white: 1, black: 0). (a) An illustrative mask with one large and several small components. (b) Its field: linear inside, logarithmic outside up to the component-aware radius of Eq.~\eqref{eqn:rscale}, which gives small components thin rings. (c) The field for one frame of a real indoor recording (ZED~2i--Livox), over the dimmed camera image. (d) The same frame with the LiDAR foreground points projected with the FGSCalib estimate; the inset enlarges the object.}
\label{fig:dt_variants}
\end{figure}

\paragraph{From binary mask to boundary-proximity field}
A binary mask provides little information to guide the search. Let $M: \Omega \to \{0, 1\}$ be the foreground mask over the pixel grid $\Omega \subset \mathbb{Z}^2$, with foreground set $\Omega_{\mathrm{fg}} = \{q \in \Omega \mid M(q) = 1\}$. A cost that counts the projected points inside $\Omega_{\mathrm{fg}}$ is the same for every projection that stays inside a region, and outside it says nothing about how close a projection is to the foreground, so the search can settle far from the correct pose. We therefore replace the mask with a continuous \emph{boundary-proximity field} $I: \Omega \to [0, 1]$ that is highest on the foreground boundary and decays on both sides (Fig.~\ref{fig:dt_variants}). The boundary is where a mask constrains the pose: a point deep inside a large region stays inside it under a shift of many pixels, whereas a point near the boundary does not. Cost fields built from distance transforms are standard in matching~\cite{borgefors1988hierarchical} and have been used for camera--LiDAR alignment on image edges~\cite{levinson2013automatic} and on semantic~\cite{autos_regi} and instance~\cite{autos_n1} masks. What differs here is the signal: class-free foreground whose component sizes vary by orders of magnitude within one frame. We therefore let the reach of the field outside a component depend on its area. Inside the boundary the profile is linear, outside it is logarithmic, although a linear outside profile performs about the same (Sec.~\ref{sec:ablation_selection}).

\paragraph{Inside profile}
Inside the foreground we use the Euclidean distance to the nearest background pixel,
\begin{equation}
\label{eqn:dt_inside}
d_{\mathrm{in}}(q) = \min_{r \in \Omega \setminus \Omega_{\mathrm{fg}}} \|q - r\|_2, \qquad q \in \Omega_{\mathrm{fg}},
\end{equation}
computed with the $3{\times}3$ chamfer approximation of Borgefors~\cite{borgefors1986distance}, normalised by its maximum $d_{\max}$ over the frame:
\begin{equation}
\label{eqn:dt_poly}
I_{\mathrm{in}}(q) = 1 - \frac{d_{\mathrm{in}}(q)}{d_{\max}}, \qquad q \in \Omega_{\mathrm{fg}}.
\end{equation}
Because $d_{\max}$ is taken over the whole frame, $I_{\mathrm{in}}$ stays close to 1 inside components much smaller than the largest one and is low only in the middle of the largest, so the cost puts its weight on the boundaries.

\paragraph{Outside profile with a component-aware attraction radius}
Outside the foreground, $d_{\mathrm{out}}(q) = \min_{r \in \Omega_{\mathrm{fg}}} \|q - r\|_2$ is the distance to the nearest foreground pixel, and the field falls from 1 at the boundary to 0 at the \emph{attraction radius}. With no limit on this radius the field would fill the image background and pull projections towards distant, unrelated regions; with one radius for all components, the small spurious components left by per-pixel background subtraction would reach as far as real objects. The radius therefore grows linearly with the area $A_k$ of component $k$, up to $\rho_{\max}$:
\begin{equation}
\label{eqn:rscale}
\rho_k = \rho_{\min} + (\rho_{\max} - \rho_{\min})\cdot\min\!\left(\frac{A_k}{A_{\mathrm{ref}}},\;1\right).
\end{equation}
This assumes a pixel scale for the apparent size of objects, but nothing about their class or shape. Each background pixel $q$ takes the radius $\rho(q)$ of its nearest component, and the outside field is, for $q\notin\Omega_{\mathrm{fg}}$,
\begin{equation}
\label{eqn:invdt_ccomp}
I_{\mathrm{out}}(q) = \left[\,1-\frac{\ln(1+\kappa\,d_{\mathrm{out}}(q))}{\ln(1+\kappa\,\rho(q))}\right]_{+},
\end{equation}
where $[x]_{+}=\max(0,x)$ and $\kappa$ sets the steepness: the field changes fastest next to the boundary and stays above zero out to the radius. Beyond the radius the field is 0, so far from the correct pose the search relies on the random redraws of the annealing (Sec.~\ref{sec:optimization}) rather than on the field. The implementation keeps the field at 1 in a band of width $\delta$ outside the boundary, raises it to an exponent $\eta$ and smooths the radius map (Table~\ref{tab:hyperparams}; full expression in the supplement, Sec.~S2).

\paragraph{The field and the alignment cost}
The field $I$ is $I_{\mathrm{in}}$ inside $\Omega_{\mathrm{fg}}$ and $I_{\mathrm{out}}$ outside it (Fig.~\ref{fig:dt_merged_ccomp}). In the reported real-data runs the switch between the two is made at every non-zero pixel of the stored camera mask, which also contains faint compression residue around the objects, so the field is 0 in a thin band there instead of following the outside profile; a single run per scenario with the switch at the binary mask, as in simulation, fails in 5 of the 72 real scenarios, against 0 to 2 for the reported runs (supplement, Sec.~S2). The cost of an extrinsic is one minus the mean field value at the projected LiDAR foreground points, averaged over a set $\mathcal{S}$ of frame pairs:
\begin{equation}
\label{eqn:cost-function}
F(R,t) = 1 - \frac{1}{|\mathcal{S}|} \sum_{s \in \mathcal{S}} \frac{1}{|P^L_s|} \sum_{p \in P^L_s} I_s\!\bigl(\operatorname{proj}(p, R, t)\bigr),
\end{equation}
where $I_s$ is the field of frame pair $s$, taken as $0$ for points that project outside the image, so moving the foreground out of view never lowers $F$. Each frame pair is normalised by its own point count so that dense foreground does not dominate. The \emph{in-image cost} $G(R,t;\mathcal{S})$ averages each frame pair only over its points that project inside the image. The search minimises $F$; $G$ scores single frame pairs in the selection and ranks repeated runs (Sec.~\ref{sec:optimization}).

\begin{table}[t]
\centering
\footnotesize
\setlength{\tabcolsep}{3pt}
\caption{Settings of the pipeline. A single value applies to both datasets; where two are given, they are for simulated and real data. The annealing constants are in the supplement (Sec.~S2).}
\label{tab:hyperparams}
\begin{tabularx}{\linewidth}{lccX}
\toprule
Symbol & Simulated & Real & Role \\
\midrule
$\rho_{\min}$, $\rho_{\max}$ & \multicolumn{2}{c}{2\,px, 25\,px} & min., max.\ attraction radius \\
$A_{\mathrm{ref}}$ & \multicolumn{2}{c}{5000\,px$^2$} & area at which $\rho_{\max}$ is reached \\
$\kappa$ & \multicolumn{2}{c}{0.2\,px$^{-1}$} & steepness of the outside profile \\
$\delta$ & \multicolumn{2}{c}{1\,px} & width of the band of value 1 \\
$\eta$ & \multicolumn{2}{c}{2} & exponent of the outside profile \\
$(\sigma_\rho,\lambda)$ & \multicolumn{2}{c}{(6\,px, 0.7)} & smoothing of the radius map \\
\midrule
$K_1$ & 100 & 200 & Stage-1 frame-pair budget \\
$K_2$ & \multicolumn{2}{c}{100} & Stage-2 frame-pair budget \\
$f$ & 0.5 & 1.0 & fraction of a budget picked greedily \\
$(\alpha,\beta,\gamma)_1$ & \multicolumn{2}{c}{$(0.4,\,0.6,\,0.0)$} & Stage-1 weights: image, depth, projection \\
$(\alpha,\beta,\gamma)_2$ & \multicolumn{2}{c}{$(0.2,\,0.4,\,0.4)$} & Stage-2 weights \\
$B$ & \multicolumn{2}{c}{5} & depth bins \\
DFPS image & mask & field & image used by the image and projection terms \\
\midrule
rotation box & $\pm\pi$ & $\pm20^\circ$ & per rotation-vector component \\
translation box & \multicolumn{2}{c}{$\pm1$\,m} & per axis \\
\bottomrule
\end{tabularx}
\end{table}
\subsection{Diversity-Aware Frame-Pair Selection (DFPS)}
\label{sec:data_selection}

Many frame pairs of a recording show the foreground at nearly the same image position and depth, and so constrain the extrinsic in nearly the same way. The search evaluates the cost hundreds of thousands of times, so each of its two stages (Sec.~\ref{sec:optimization}) works on a subset of the candidate pool $\mathcal{C}=\{c_i\}_{i=1}^{N}$. The subset has $K_1$ frame pairs in Stage~1 and $K_2$ in Stage~2 (Table~\ref{tab:hyperparams}), or the whole pool when it is smaller. DFPS spends this budget on frame pairs that differ in where and at what depth the foreground appears. It initialises $\mathcal{S}$ as an empty set and iteratively adds the highest-scoring candidate among those not yet selected, with the score
\begin{equation}
Q(c_i \mid \mathcal{S})=\alpha\,\widehat{D}_{\mathrm{img}}(c_i, \mathcal{S})+\beta\,\widehat{D}_{\mathrm{depth}}(c_i, \mathcal{S})+\gamma\,\widehat{C}_{\mathrm{proj}}(c_i),
\label{eq:selection_score_compact}
\end{equation}
until a fraction $f$ of the budget is filled; the remaining part of the budget is drawn uniformly at random from the candidates not yet selected. The weights satisfy $\alpha+\beta+\gamma=1$, and the hat notation ($\widehat{D}$, $\widehat{C}$) denotes min--max normalisation over the remaining candidates.

\emph{Image diversity} $D_{\mathrm{img}}$ is the mean distance from the centroid of the non-zero pixels of the candidate's image (the mask in simulation, the field on real data) to the centroids of the frame pairs already selected. \emph{Depth diversity} $D_{\mathrm{depth}}$ is 1 if the mean foreground depth of the candidate falls in one of $B$ equal-width depth bins not yet covered by $\mathcal{S}$, and 0 otherwise; objects at different distances constrain rotation and translation differently. \emph{Projection consistency} $C_{\mathrm{proj}}(c_i)=1-G(R_1,t_1;\{c_i\})$ is used only in Stage~2; it favours frame pairs on which the Stage-1 estimate $(R_1,t_1)$ already has a low in-image cost. The weights and $f{=}0.5$ were set on the simulated data, where drawing half of each budget at random kept one scene from converging to an incorrect estimate; $K_1$ and $f{=}1$ were set on the real recordings (supplement, Sec.~S1).

\subsection{Two-Stage Simulated-Annealing Search}
\label{sec:optimization}

$F$ uses discrete pixel lookups, so it is not differentiable, and it has many local minima. Methods that refine a near-correct calibration, such as~\cite{autos_regi}, can search locally; for a wider search we use simulated annealing (SA)~\cite{kirkpatrick1983sa}, a random search that also accepts some worsening moves, less often as it proceeds, so that it can leave local minima. Algorithm~\ref{alg:pipeline} combines it with DFPS in two stages.

\begin{algorithm}[t]
\caption{FGSCalib extrinsic calibration}
\label{alg:pipeline}
\footnotesize
\begin{algorithmic}[1]
\Require sequence of frame pairs; camera intrinsics; nominal pose and search box; budgets $K_1,K_2$; stage weights $(\alpha,\beta,\gamma)_{1,2}$; greedy fraction $f$
\Ensure extrinsic $(R^*,t^*)$ and its in-image cost $G$
\For{each frame pair $s$}\Comment{segmentation, Sec.~\ref{sec:Featureextraction}}
  \State $M_s \gets \textsc{MoG}_{\mathrm{cam}}(s)$;\quad $P^L_s \gets \textsc{MoG}_{\mathrm{lidar}}(s)$
  \State $I_s \gets \textsc{BuildField}(M_s)$ \Comment{boundary-proximity field, Sec.~\ref{sec:dt_invdt}}
\EndFor
\State $\mathcal{C} \gets \{(I_s,P^L_s):|P^L_s|>0\}$ \Comment{candidate pool}
\Statex \textbf{Stage 1: broad search} (no estimate yet)
\State $\mathcal{S}_1 \gets \textsc{DFPS}(\mathcal{C}, K_1, (\alpha,\beta,\gamma)_1, \varnothing, f)$ \Comment{Sec.~\ref{sec:data_selection}}
\State $(R_1,t_1) \gets \textsc{SA}(F \text{ over } \mathcal{S}_1;\ \text{start}=\text{nominal pose})$
\Statex \textbf{Stage 2: refinement} (projection term evaluated at $(R_1,t_1)$)
\State $\mathcal{S}_2 \gets \textsc{DFPS}(\mathcal{C}, K_2, (\alpha,\beta,\gamma)_2, (R_1,t_1), f)$
\State $(R^*,t^*) \gets \textsc{SA}(F \text{ over } \mathcal{S}_2;\ \text{start}=(R_1,t_1))$
\State \Return $(R^*,t^*)$ and $G(R^*,t^*)$ on $\mathcal{S}_2$
\Statex \emph{Real data:} five runs of lines 6--10; the estimate with the lowest $G$ is kept.
\end{algorithmic}
\end{algorithm}

\paragraph{Search space and start}
Rotation is parameterised as an axis-angle (Rodrigues) vector, and translation is searched within $\pm1$\,m per axis around zero. The nominal pose has zero translation and the rotation that only converts the LiDAR axis convention ($x$ forward, $y$ left, $z$ up) into the camera's ($x$ right, $y$ down, $z$ forward). In simulation each component of the rotation vector is searched over $[-\pi,\pi]$, which covers every rotation. On the real rig, whose sensors are mounted facing roughly the same way, each component is searched within $\pm0.35$\,rad ($20^{\circ}$) of its nominal value; as the nominal rotation vector has a norm of $120^\circ$, this box is not a ball of rotations about the nominal pose. It contains every reference with a margin of at least $3.9^\circ$ per component (supplement, Sec.~S2).

\paragraph{Two stages}
Stage~1 starts at the nominal pose and minimises $F$ over $K_1$ frame pairs selected with weights that favour the depth term and switch off the projection term, as no estimate exists yet. Stage~2 re-selects $K_2$ frame pairs with weights that include the projection term at the Stage-1 estimate $(R_1,t_1)$, and searches again from it, with the same box and annealing schedule.

\paragraph{Annealing and repeated runs}
With probability 0.9 an SA proposal composes the current rotation with a Gaussian axis-angle rotation and perturbs the translation additively; otherwise all six parameters are redrawn uniformly from the box. Amplitude and temperature decay geometrically under Metropolis acceptance, with one schedule for every experiment; seven chains run in parallel and the best state found is returned. A \emph{run} is one execution of the two stages on one scenario (one sensor pair in one recording, Sec.~\ref{sec:dataset}); as the search is random, runs with different seeds can end at different estimates. On the real data every scenario is run five times and the estimate with the lowest final in-image cost $G$ on its own Stage-2 frame pairs is kept, a rule that needs no reference. Simulated results are single runs unless stated otherwise.

\section{Dataset and Experimental Setup}
\label{sec:dataset}

\begin{figure*}[!t]
    \centering
    \includegraphics[width=0.94\textwidth]{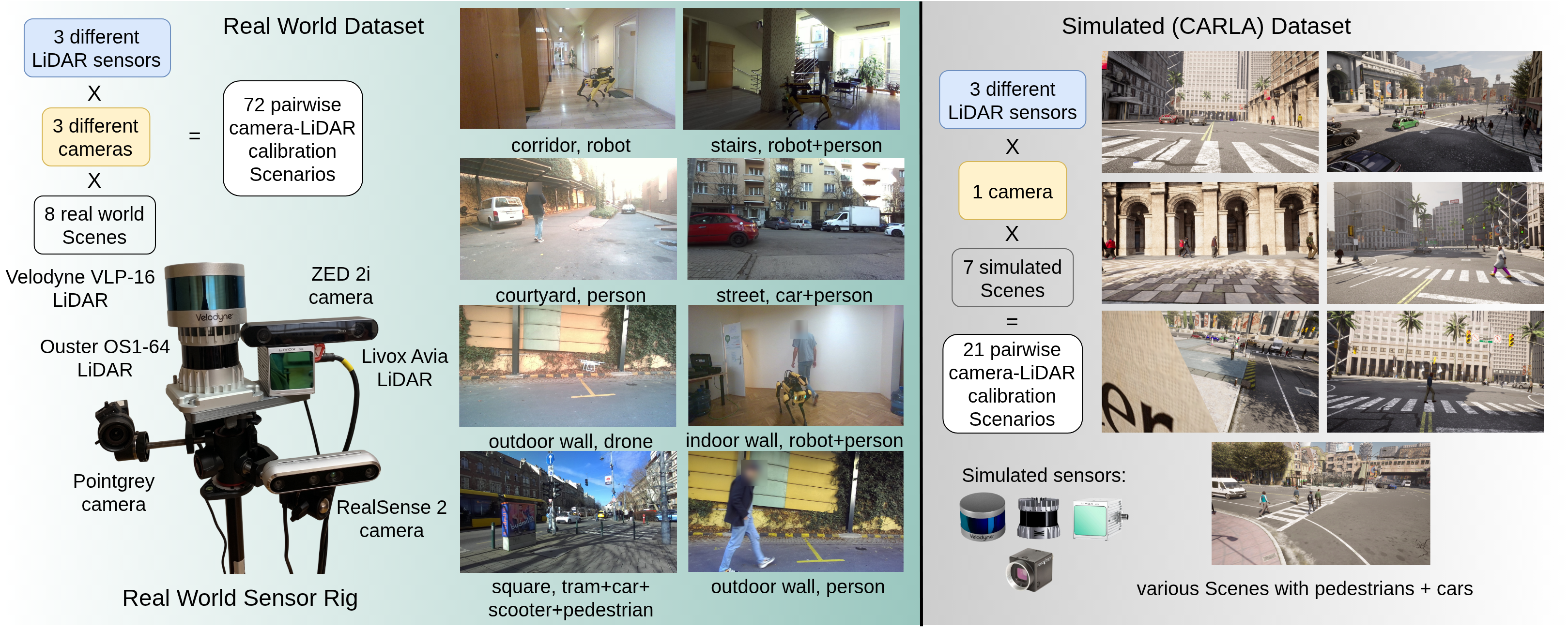}
    \caption{Dataset overview. \textbf{Left:} real sensor rig (Velodyne VLP-16, Ouster OS1-64, Livox Avia LiDARs; Pointgrey, RealSense~2, ZED~2i cameras) and representative images from the 8 scenes, yielding 72 scenarios (3 LiDARs $\times$ 3 cameras $\times$ 8 scenes). Foreground objects span pedestrians, vehicles, trams, a drone, and a quadruped robot. \textbf{Right:} CARLA data with three simulated LiDARs and one camera across 7 scenes, yielding 21 scenarios with exact extrinsics.}
    \label{fig:dataset_overview}
\end{figure*}

Driving datasets such as KITTI~\cite{geiger2013vision} record from a moving platform, the complementary case of a rigid world and a moving sensor, and infrastructure datasets such as DAIR-V2X-I~\cite{dairv2x} provide one camera--LiDAR configuration with a single extrinsic and no repeated calibration sessions. We therefore built a benchmark of CARLA simulations and real multi-sensor recordings (Fig.~\ref{fig:dataset_overview}), designed for variety rather than scale. A \emph{sensor pair} is one camera with one LiDAR, a \emph{scene} is one recording at one place, and a \emph{scenario} is one sensor pair in one scene.

\subsection{Simulated Dataset}
We place virtual sensors in the CARLA simulator~\cite{dosovitskiy2017carla} and derive the camera--LiDAR extrinsic from their known poses, so the simulated ground truth is exact; the nominal start is 19 to 61\,cm and $3.4^\circ$ to $6.7^\circ$ from it. We simulate a 16-channel and a 64-channel RMB LiDAR (labelled 16-channel and 64-channel where the tables split them) and an NRCS LiDAR with the rosette pattern of the Livox Avia, and record 600 frames at 10\,Hz at each of 7 fixed sensor placements in CARLA's Town10HD map with moving pedestrians and vehicles.

\subsection{Real Dataset}
We recorded 8 scenes with one rig: a narrow indoor corridor, an open courtyard, urban streets, a busy square, a staircase, and plain-wall backgrounds. The foreground objects (pedestrians, vehicles, trams, a quadruped robot and a drone) are deliberately varied, since the method uses no object class. The scenes range from an easy one, a staircase with a person and a robot, to deliberately hard ones: plain walls with little static structure, a single small drone, a short recording of a crowded square with trams and cars, and a street with cars. Three LiDARs and three cameras on a stationary platform, not reassembled between recordings, give nine sensor pairs per scene and 72 scenarios. All three LiDARs deliver 10 sweeps per second, so a sweep of the NRCS LiDAR integrates 100\,ms. We use only one RGB stream of each camera, with no stereo or depth; a scenario provides a few hundred candidate frame pairs, fewer in the short recording of the square. Four further recordings with the same rig contain no moving objects; FGSCalib cannot use them, and they serve only to run the static-scene baselines in their own conditions.

\subsection{Reference Calibration and Its Uncertainty}
\label{sec:reference}
The real recordings have no exact ground truth. Our \emph{reference} is the average of independent manual calibrations: 43 sessions (3--6 per sensor pair) with the Autoware toolkit~\cite{kato2018autoware}, each from a different recording: seven with a calibration board and 36 with correspondences picked by hand on natural features in the scene recordings. The sessions of a sensor pair deviate from their mean by an RMS deviation $s$ (Euclidean in translation, geodesic in rotation, $n{-}1$ degrees of freedom) whose median over the nine sensor pairs is 9.4\,cm and $0.88^\circ$ (supplement, Sec.~S10); as the sessions use different recordings, $s$ describes reproducibility rather than repeatability~\cite{vim2012}. The Type~A standard uncertainty of the session mean~\cite{jcgm100gum}, $s/\sqrt{n}$, is the \emph{reference uncertainty} $u_t$, $u_R$; its medians are 4.0\,cm and $0.39^\circ$, and on Pointgrey--Velodyne $u_R$ reaches $1.0^\circ$. It excludes systematic effects shared by all sessions, such as the camera intrinsics.

\subsection{Evaluation Protocol}
\label{sec:evaluation}
An estimate is compared with the exact simulator pose or with the real-data reference: its error $\Delta t$ is the distance between the two translations and $\Delta R$ the angle of the relative rotation. On real data the error therefore includes the uncertainty of the reference.

\paragraph{Tolerance from manual calibration} An estimate is \emph{within tolerance} when it is within 18\,cm and $1.8^\circ$, both bounds required. The bound comes from the manual sessions. We compare each of the 43 sessions with the mean of the other sessions of its sensor pair; 95\% of these differences are within 17.9\,cm and 95\% within $1.81^\circ$, which we round to 18\,cm and $1.8^\circ$. At the rounded bound 40 of the 43 sessions are within both, so a single manual calibration meets the tolerance in 93\% of cases by construction, and an estimate within it is about as close to the reference as a single manual calibration is to the others. The tolerance thus reflects what the reference can resolve, not the needs of an application; with our cameras $1.8^\circ$ moves a projected point by 17 to 30\,px at any depth. We report the same bound in simulation. The comparison does not hinge on this tolerance: for every pair of bounds up to 50\,cm\,/\,$5^\circ$ with a translation bound of at least 2\,cm and a rotation bound above $0.3^\circ$, the lowest-cost FGSCalib run is within them in more scene-recording scenarios than any baseline (supplement, Sec.~S9).

\paragraph{Failure} An estimate more than 50\,cm or $5^\circ$ off is a \emph{failure}, a gross error.

\section{Results}
\label{sec:experiments}

We run every method on the recordings with moving objects, and the two methods designed for static scenes also on static recordings. The comparison therefore shows where each method's assumptions hold, not a general ranking.

\begin{figure}[t]
\centering
\includegraphics[width=\linewidth]{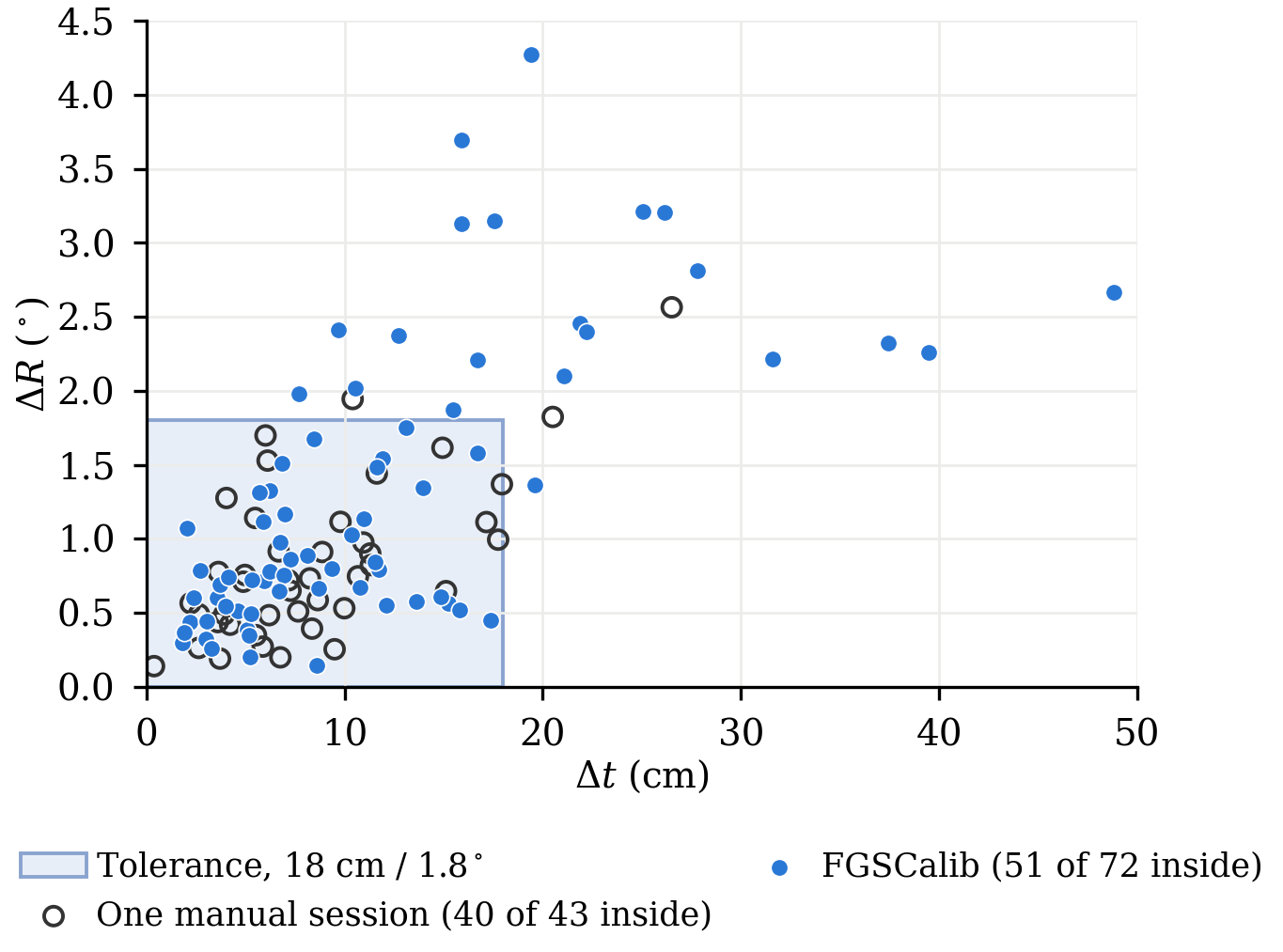}
\caption{Error of each real FGSCalib estimate (72 scenarios, lowest-cost run of five) and of each single manual calibration against the other sessions of its sensor pair (43 sessions). The tolerance box rounds the 95th percentiles of the manual differences, so 40 of the 43 manual sessions lie inside it by construction.}
\label{fig:cdf}
\end{figure}

\begin{table*}[t]
\centering
\footnotesize
\setlength{\tabcolsep}{2.8pt}
\caption{Real data, grouped by condition. Scene recordings with moving objects: 24 scenarios per LiDAR (8 scenes $\times$ 3 cameras). Static recordings without moving objects: Livox only (4 recordings $\times$ 3 cameras; with the RMB LiDARs Koide et al.\ fails in all 24 static runs). Counts are of 24 unless stated: Koide et al.\ returned no estimate for one Ouster scenario, and Yuan et al.\ gives one joint calibration per sensor pair (3), or ten per pair from coarse starts within $\pm5^\circ$, $\pm10$\,cm of the reference (30). Each cell: estimates within tolerance (18\,cm\,/\,$1.8^\circ$) and median error $\Delta t$ (cm)\,/\,$\Delta R$ ($^\circ$) over all estimates; $>$50\,/\,$>$5: beyond the failure bound. Bold: most estimates within tolerance among the methods.}
\label{tab:real_combined}
\begin{tabularx}{\textwidth}{@{}l>{\raggedright\arraybackslash}Xlrrrrrr@{\hspace{8pt}}rr@{}}
\toprule
& & & \multicolumn{6}{c}{Scene recordings, with moving objects} & \multicolumn{2}{c}{Static recordings} \\
\cmidrule(lr){4-9}\cmidrule(l){10-11}
& & & \multicolumn{2}{c}{Ouster} & \multicolumn{2}{c}{Velodyne} & \multicolumn{2}{c}{Livox} & \multicolumn{2}{c}{Livox} \\
\cmidrule(lr){4-5}\cmidrule(lr){6-7}\cmidrule(lr){8-9}\cmidrule(l){10-11}
Method & Requirements (from its paper) & Start & in tol. & median & in tol. & median & in tol. & median & in tol. & median \\
\midrule
\multicolumn{11}{@{}l}{\textbf{\emph{Learned or segment-based}}} \\[1pt]
LCCNet~\cite{lccnet} & trained on one rig (KITTI); start within 1.5\,m, $20^\circ$ & nominal & 0 & $>$50/$>$5 & 0 & $>$50/$>$5 & 0 & $>$50/$>$5 & \multicolumn{2}{c}{not run} \\
CMRNext~\cite{cmrnext} & trained on driving data; start within 2\,m, $10^\circ$ & nominal & 0 & $>$50/$>$5 & 1 & $>$50/$>$5 & 0 & $>$50/$>$5 & \multicolumn{2}{c}{not run} \\
Calib-Anything~\cite{calibanything} & start within $10^\circ$, 0.5\,m; intensity, normals & nominal & 0 & $>$50/$>$5 & 0 & $>$50/$>$5 & 2 & $>$50/$>$5 & \multicolumn{2}{c}{not run} \\
\midrule
\multicolumn{11}{@{}l}{\textbf{\emph{Static-scene features}}} \\[1pt]
Koide et al.~\cite{koide2023general} & static scene; dense cloud with intensity & own guess & 0 of 23 & $>$50/$>$5 & 0 & $>$50/$>$5 & 11 & 13.9/1.59 & \best{11 of 12} & 4.7/0.53 \\
Yuan et al.~\cite{hku_calib} & static planar edges; dense cloud; initial pose & nominal & \multicolumn{4}{c}{not run (needs a dense cloud)} & 0 of 3 & 30.2/$>$5 & \multicolumn{2}{c}{--} \\
Yuan et al. & as above & coarse & \multicolumn{4}{c}{not run (needs a dense cloud)} & 6 of 30 & 15.6/4.19 & 1 of 3 & 18.7/0.90 \\
\midrule
\multicolumn{11}{@{}l}{\textbf{\emph{Moving objects}}} \\[1pt]
FGSCalib (ours), one run & moving objects; rotation $\pm20^\circ$ & nominal & 11 & 15.0/1.66 & 17 & 11.2/1.11 & 18 & 7.5/0.67 & \multicolumn{2}{c}{\multirow{2}{*}{\shortstack{cannot run:\\no moving objects}}} \\
\quad lowest-cost run of five & as above & nominal & \best{13} & 12.6/1.46 & \best{18} & 10.9/1.23 & \best{20} & 6.8/0.74 & & \\
\midrule
One manual session & an operator; vs.\ the others & -- & 13 of 13 & 8.9/0.91 & 10 of 13 & 10.0/0.75 & 17 of 17 & 6.0/0.49 & \multicolumn{2}{c}{--} \\
\bottomrule
\end{tabularx}
\end{table*}

\subsection{FGSCalib against Manual Calibration}
\label{sec:manual}

\emph{Most FGSCalib estimates are as close to the reference as single manual calibrations are} (Fig.~\ref{fig:cdf}). Keeping the lowest-cost of five runs, a rule that needs no reference, FGSCalib is within tolerance in 51 of the 72 scenarios, with all three LiDARs, and none of the 72 estimates is a failure (Table~\ref{tab:real_combined}). Its errors are somewhat larger than those of a single manual session, mostly in rotation: 18 of its 72 estimates are more than $2^\circ$ off, against one of the 43 sessions. In simulation, with exact ground truth, it is within tolerance in 81 of 84 runs (Table~\ref{tab:ablation_main}(a), row~5).

\emph{The start of the search matters little.} A single run from a random pose anywhere in the search box, up to $35^\circ$ and 1.55\,m from the reference, is within tolerance about as often as a single run from the nominal pose (47 against 46 of 72 scenarios, with 3 and 2 failures; Table~\ref{tab:real_pair} and supplement, Sec.~S7). In simulation, where the box covers every rotation, it also converges from starts $59^\circ$ to $176^\circ$ away (supplement, Sec.~S7).

\begin{figure*}[t]
\centering
\makebox[0.217\linewidth][c]{\small Start (nominal pose)}%
\makebox[0.217\linewidth][c]{\small FGSCalib (ours)}%
\makebox[0.217\linewidth][c]{\small Ground truth\,/\,reference}\\[2pt]
\includegraphics[width=0.65\linewidth]{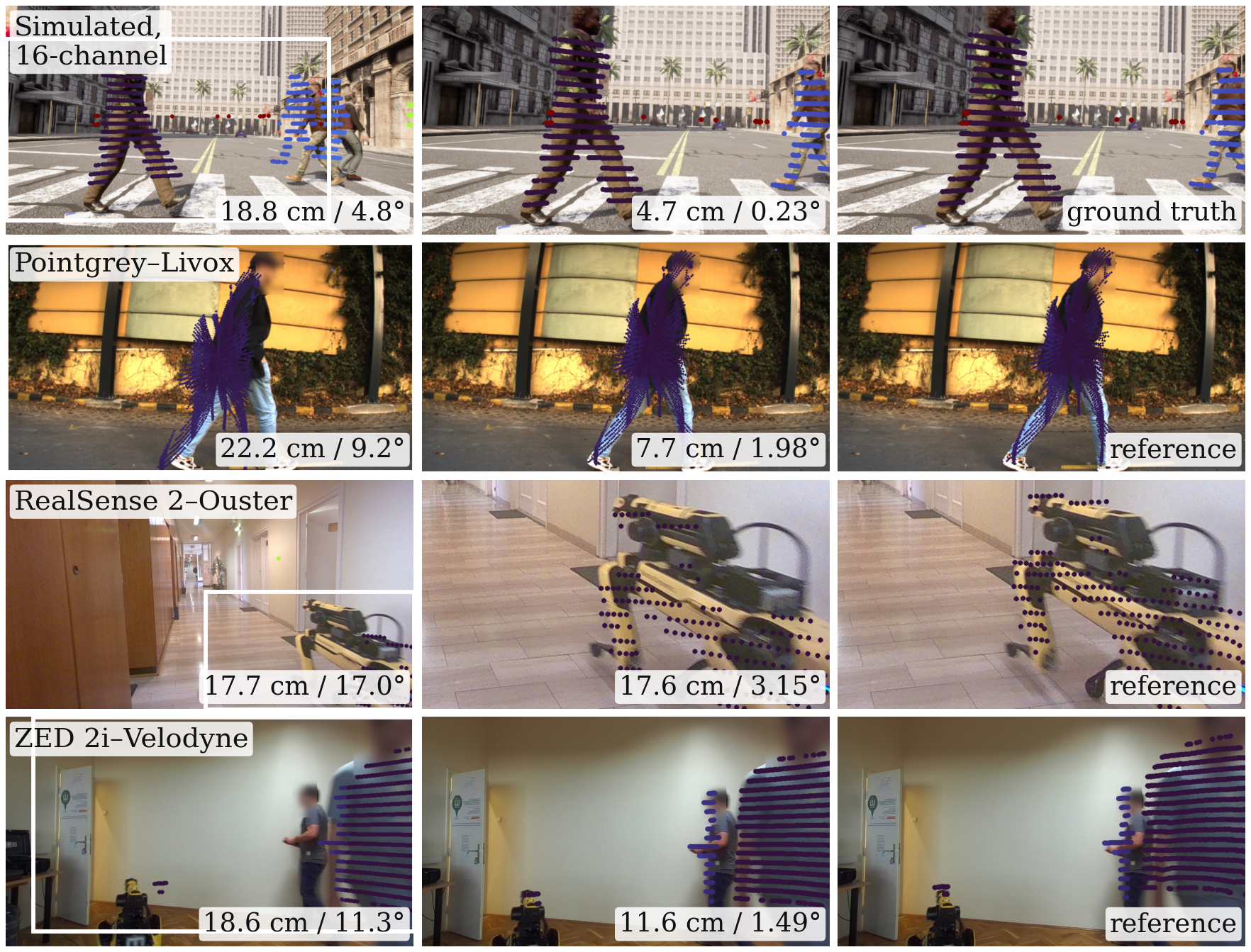}
\caption{LiDAR foreground points, coloured by depth, projected into the camera image. Column~1: the whole image at the start of the search (nominal pose); the white frame marks the region that columns 2 and 3 enlarge. Columns 2 and 3: the FGSCalib estimate (on the real rows the lowest-cost run of five) and the ground truth (row~1) or the reference (rows 2--4). Each cell gives its $\Delta t$\,/\,$\Delta R$. Rows 2--4: for three sensor pairs, the scenario with the median $\Delta t$ of the pair (one of the two middle ones), on the frame with the most LiDAR foreground points: an outdoor wall, a corridor and an indoor wall. Rows 2 and 3 are outside the rotation tolerance although the overlay looks close; row 4 is within the tolerance. Row 2 uses smaller dots, as the Livox gives a dense cloud.}
\label{fig:projection}
\end{figure*}

\emph{Accuracy depends on the scene.} FGSCalib is within tolerance with all nine sensor pairs in the staircase, in 6 of 9 with a single drone as the only moving object, and in 4 of 9 in the street. Cars move fastest there, and for them the time offset between image and sweep likely matters most (Sec.~\ref{sec:discussion}). On the two plain-wall scenes, where static-scene methods find little structure, it is within tolerance in 13 of 18 scenarios. The weakest sensor pairs are the Ouster with the RealSense~2 and with the ZED~2i (Table~\ref{tab:real_pair}). Fig.~\ref{fig:projection} shows the median scenario of three sensor pairs.

\subsection{Comparison with the Baselines}
\label{sec:comparison}

Each baseline runs in its recommended configuration on the same recordings. Koide et al.\ computes its own initial guess; the other baselines start, like FGSCalib, from the nominal pose (Yuan et al.\ also from coarse starts), which lies 12 to 28\,cm and $1.2^\circ$ to $19^\circ$ from the reference and, by chance, inside the tolerance for ZED~2i--Livox. LCCNet is trained for such errors, while CMRNext and Calib-Anything expect at most $10^\circ$, which the nominal pose exceeds for some sensor pairs. Our recordings lie outside the training domain of the learned methods, so Table~\ref{tab:real_combined} tests them only there; their accuracy within that domain is reported in their own papers.

\emph{With the Livox, Koide et al.\ also reaches the tolerance in many scenarios among moving objects, and Yuan et al.\ in some from a coarse start; with the RMB LiDARs only FGSCalib does} (Table~\ref{tab:real_combined}). The Livox's NRCS pattern fills in a dense image of the static background over time. With it Koide et al.'s median error is inside the tolerance, and FGSCalib is within tolerance in more scenarios (20 against 11 of 24); Yuan et al., started within $\pm5^\circ$ and $\pm10$\,cm of the reference (a coarse start, which uses the reference), is within it in 6 of 30 runs. A stationary RMB LiDAR gives only a sparse point cloud, and Koide et al.\ fails with it on the static recordings as well (24 of 24), which points to the sparse cloud, not the moving objects, as the cause. With these LiDARs FGSCalib is within tolerance in 31 of 48 scenarios, and of all baseline estimates only one is. The released learned models did not transfer to this rig; LCCNet fails in every scenario.

\emph{On the static recordings the roles reverse.} FGSCalib cannot run, because nothing moves, while Koide et al.\ is within tolerance in 11 of 12 calibrations, close to the accuracy its authors report (supplement, Sec.~S3). Yuan et al., from a coarse start, is within tolerance in one of its three calibrations there and close to it in the other two. The methods therefore cover complementary situations: static structure seen by a dense LiDAR, or moving objects seen by rotating and non-repetitive LiDARs alike.

In simulation no baseline is within tolerance in more than 2 of 21 scenarios (supplement, Sec.~S11). Koide et al.\ is not run there, because the LiDAR intensity of our CARLA data depends only on range, not on the surface. The simulated scenes fit the assumptions of FGSCalib, whose selection weights were also set on them, so that comparison favours FGSCalib and the real data are the primary test.

\begin{table}[t]
\centering
\footnotesize
\setlength{\tabcolsep}{3pt}
\caption{FGSCalib per sensor pair: median error of the lowest-cost run of five ($\Delta t$ in cm, $\Delta R$ in degrees), scenes within tolerance (of 8) for the lowest-cost run, for one run from the nominal pose and for one run from a random start, and the reference uncertainty $u_t$ (cm), $u_R$ ($^\circ$) (Sec.~\ref{sec:reference}).}
\label{tab:real_pair}
\begin{tabular}{lrrrrrrr}
\toprule
& \multicolumn{2}{c}{Median} & \multicolumn{3}{c}{Within tolerance} & \multicolumn{2}{c}{Ref.\ unc.} \\
\cmidrule(lr){2-3}\cmidrule(lr){4-6}\cmidrule(lr){7-8}
Sensor pair & $\Delta t$ & $\Delta R$ & low.\ cost & 1 run & rand. & $u_t$ & $u_R$ \\
\midrule
Pointgrey--Livox & 7.3 & 0.78 & 7 & 6 & 7 & 3.3 & 0.47 \\
RealSense~2--Livox & 6.3 & 0.83 & 7 & 6 & 6 & 2.2 & 0.23 \\
ZED~2i--Livox & 8.5 & 0.63 & 6 & 6 & 6 & 2.1 & 0.14 \\
\midrule
Pointgrey--Ouster & 6.8 & 0.72 & 7 & 6 & 7 & 4.8 & 0.49 \\
RealSense~2--Ouster & 16.7 & 2.36 & 3 & 3 & 3 & 4.9 & 0.46 \\
ZED~2i--Ouster & 16.1 & 2.04 & 3 & 2 & 1 & 4.2 & 0.39 \\
\midrule
Pointgrey--Velodyne & 11.4 & 0.90 & 7 & 7 & 7 & 9.6 & 1.01 \\
RealSense~2--Velodyne & 8.6 & 1.21 & 6 & 6 & 6 & 3.1 & 0.25 \\
ZED~2i--Velodyne & 14.2 & 1.53 & 5 & 4 & 4 & 4.0 & 0.28 \\
\bottomrule
\end{tabular}
\end{table}

\subsection{Sensor Pairs and Repeated Runs}
\label{sec:agreement}

Table~\ref{tab:real_pair} gives the results per sensor pair. Along the viewing axis FGSCalib places the LiDAR farther from the camera than the reference does (66 of 72 scenarios, median 3.9\,cm), as the minimum of the simulated cost does (Sec.~\ref{sec:ablation_selection}). The supplement (Sec.~S6) adds checks that need no reference: agreement across scenes, and consistency around loops of two cameras and two LiDARs.

\emph{One run does not settle whether a scenario is within tolerance.} The five runs of a scenario usually agree closely (their spread has a median of 2.7\,cm and $0.34^\circ$), yet in 20 of the 72 scenarios some are within tolerance and others are not. Keeping the run with the lowest cost avoided every failed run in our data (7 of 360) but does not reliably pick the most accurate run (supplement, Sec.~S4). A run takes a median of 11\,min on one desktop CPU (Intel i7-7700K), so five runs take about an hour; a simulated run takes 26\,min because its foreground point sets are denser.

\begin{table*}[t]
\centering
\footnotesize
\setlength{\tabcolsep}{4pt}
\begin{threeparttable}
\caption{Ablation. (a) The 21 simulated scenarios, four seeds each (84 runs per row; $K_1{=}K_2{=}100$, $f{=}0.5$; no run fails). Image: the image the cost is computed on. (b) The 72 real scenarios with the real-data settings; a repeat is one run of every scenario, and rows 3--5 are compared on the same three repeats (all five repeats of row~5: 233 of 360 runs within tolerance, 7 failures). On real data the mask also changes the frame pairs that DFPS selects (Table~\ref{tab:hyperparams}). Medians, means, standard deviations ($\sigma$) and counts are over all runs of a row; within tolerance: 18\,cm\,/\,$1.8^\circ$.}
\label{tab:ablation_main}
\begin{tabular}{rllcrrrrrrrr}
\toprule
& & & & & & \multicolumn{2}{c}{$\Delta t$ (cm)} & \multicolumn{2}{c}{$\Delta R$ ($^\circ$)} & \multicolumn{2}{c}{Max} \\
\cmidrule(lr){7-8}\cmidrule(lr){9-10}\cmidrule(lr){11-12}
\multicolumn{4}{l}{\emph{(a) Simulated}} & In tol. & $\Delta t>20$\,cm & Med. & Mean$\pm\sigma$ & Med. & Mean$\pm\sigma$ & $\Delta t$ & $\Delta R$ \\
& Selection & Image & Stages & (of 84) & (of 84) & & & & & (cm) & ($^\circ$) \\
\midrule
1 & random & mask & 1 & 76 & 4 & 5.8 & 8.1$\pm$6.2 & 0.18 & 0.25$\pm$0.25 & 25.0 & 1.53 \\
2 & random & mask & 2 & 78 & 3 & 6.6 & 7.7$\pm$5.3 & 0.18 & 0.25$\pm$0.26 & 25.0 & 1.51 \\
3 & DFPS & mask & 2 & 74 & 9 & 5.9 & 9.2$\pm$8.7 & 0.16 & 0.25$\pm$0.22 & 39.4 & 0.99 \\
4 & random & field & 2 & 83 & 0 & 5.6 & 6.4$\pm$3.7 & 0.17 & 0.22$\pm$0.16 & 19.4 & 1.09 \\
5 & DFPS & field (ours) & 2 & 81 & 2 & 5.3 & 6.9$\pm$4.9 & 0.19 & 0.21$\pm$0.11 & 27.8 & 0.58 \\
\midrule
\multicolumn{4}{l}{\emph{(b) Real}} & In tol. & Failures & & & & & & \\
& Selection & Image & Repeats & / runs & / runs & & & & & & \\
\midrule
3 & DFPS & mask & 3 & 149\,/\,216 & 10\,/\,216 & 9.3 & 15.0$\pm$18.7 & 0.87 & 1.23$\pm$1.04 & 142 & 8.05 \\
4 & random & field & 3 & 131\,/\,216 & 5\,/\,216 & 9.8 & 12.6$\pm$9.9 & 1.20 & 1.54$\pm$1.42 & 73 & 10.11 \\
5 & DFPS & field (ours) & 3 & 137\,/\,216 & 5\,/\,216 & 9.7 & 13.3$\pm$10.4 & 1.03 & 1.44$\pm$1.19 & 56 & 8.08 \\
\bottomrule
\end{tabular}
\begin{tablenotes}[para]\footnotesize
\item[] A failure is a run more than 50\,cm or $5^\circ$ off; means include the failures.
\end{tablenotes}
\end{threeparttable}
\end{table*}

\section{Ablation Study}
\label{sec:ablation_selection}

Table~\ref{tab:ablation_main} shows which parts of FGSCalib produce these results. Repeating a configuration with another seed changes the error of a simulated scenario by a median of 0.9\,cm\,/\,$0.05^\circ$ and the median of a row of Table~\ref{tab:ablation_main}(a) by 0.5 to 2.1\,cm, so differences of that size are not resolved by this experiment. Four further variants of the field (no limit on the attraction radius, one radius for all components, the inside profile alone, and a linear outside profile) are in the supplement (Sec.~S5); none is clearly better than ours, and the linear and logarithmic outside profiles cannot be told apart.

\emph{The field and the binary mask reach the same median error; the mask has more large errors.} In simulation the medians of the five rows differ by about the rerun variation. Under either selection rule the mask has more runs above 20\,cm than the field, 9 against 2 with DFPS (eight of the nine in the two RMB-LiDAR scenarios of one scene) and 3 against none with random selection, and fewer runs within tolerance; with the field the mean $\Delta t$ is lower and its spread smaller. On real data the mask is within tolerance slightly more often and has a lower median $\Delta R$, but it fails twice as often and its largest error is more than twice as large (Table~\ref{tab:ablation_main}(b)).

\emph{The remaining simulated error is mainly a bias of the cost.} Evaluated on their own frame pairs, all 84 runs of row~3 and all 84 of row~5 end at a lower cost $F$ than the exact ground truth. A local search around ground truth (supplement, Sec.~S8) finds a minimum of $F$ a median of 5.8\,cm\,/\,$0.19^\circ$ from ground truth with the field and 4.7\,cm\,/\,$0.12^\circ$ with the mask, offset away from the camera along the viewing axis in 19 and 16 of the 21 scenarios and downwards in the image; the two minima differ in no consistent direction. The runs with the field end a median of 1.8\,cm from this minimum, about the resolution of the check, so the search finds the minimum and the cost places it off the ground truth.

\emph{The second stage corrects some failures; DFPS lowers the largest rotation errors and the spread of the rotation error.} On real data the second stage corrects four of the eleven Stage-1 failures in our 360 runs and introduces none, while the number within tolerance (231 before, 233 after) and the median $\Delta t$ (9.3 and 9.7\,cm) hardly change. In simulation, where no run fails, the median $\Delta t$ of row~5 falls from the Stage-1 to the final estimate from 6.6 to 5.3\,cm, although the 95\% interval of this change, from resampling the seven scenes, is $-1.9$ to $+0.8$\,cm. With the field, DFPS and random selection reach about the same median error (rows 4 and 5); the benefit of DFPS is in the rotation. In simulation it halves the largest rotation error ($0.58^\circ$ against $1.09^\circ$) and narrows the rotation spread ($\sigma$ of $0.11^\circ$ against $0.16^\circ$), while the counts within tolerance differ by two runs of 84. On real data, over the same three repeats, its largest errors are 56\,cm\,/\,$8.1^\circ$ against 73\,cm\,/\,$10.1^\circ$ and its rotation spread $1.19^\circ$ against $1.42^\circ$, with 137 against 131 of 216 runs within tolerance and 5 failures each. The number of runs above 20\,cm is not reduced (supplement, Sec.~S5).

\section{Discussion}
\label{sec:discussion}

\paragraph{Why the signal works}
Moving objects give a stationary rig what it otherwise lacks: correspondences that change over time and cover different image regions and depths, and that even a sparse RMB LiDAR samples, because the objects cross its rings. This is consistent with FGSCalib reaching the tolerance in most scenarios with each of the three LiDARs, while the static-scene baselines are within tolerance only with the Livox, whose NRCS pattern fills in over time. The two kinds of cue are complementary (Sec.~\ref{sec:comparison}).

\paragraph{What carries the accuracy}
The signal and the global search set the median accuracy, which a plain binary mask already reaches. The field and the second stage reduced gross errors in our tests: with the field the largest real-data error fell from 142 to 56\,cm and the simulated runs above 20\,cm from 9 to 2 of 84, and the second stage corrected four of the eleven Stage-1 failures on real data without introducing any. For an unattended installation, where a gross error would go unnoticed, this matters more than the median, but the method has no reliable reference-free failure detector yet: the lowest-cost rule avoided every failed run in our data, while among the others $G$ is a weak guide. What remains in simulation is a bias of the cost of a few centimetres, away from the camera and downwards, and the real estimates are offset away from the camera too; this points to the next step, a cost that also penalises camera foreground left uncovered.

\paragraph{Scope and limitations}
All real recordings come from one tripod-mounted rig, with the rotation searched in a $\pm20^\circ$ box. We did not evaluate pole- or gantry-mounted installations, which see objects from farther away, nor remounting; recovery from arbitrary rotations is shown only in simulation (supplement, Sec.~S7). Frame pairs were formed by timestamp without hardware triggering: the image and the sweep are a median of 11\,ms and at most 103\,ms apart, the Livox and Ouster stamps are reception times, and object motion within a sweep is not compensated, which matters most for fast objects such as cars. Some settings were chosen on the real recordings and others on the simulated ones (supplement, Sec.~S1), so neither evaluation is fully held out.

\section{Conclusion}

FGSCalib calibrates a stationary camera--LiDAR pair from whatever moves through its view, with no calibration target, no platform motion and no measured initial extrinsic, and it works with both RMB and NRCS LiDARs. Keeping the lowest-cost of five runs, it is within the tolerance that 93\% of single manual calibrations meet (18\,cm\,/\,$1.8^\circ$) in 51 of 72 real scenarios, with all three LiDARs: in 20 of 24 with the Livox, where the best baseline reaches 11, and in 31 of 48 with the RMB LiDARs, where only one baseline estimate does. On static recordings, where FGSCalib cannot run, a static-scene method with a dense LiDAR is the right choice. Its median error is 9.5\,cm\,/\,$0.9^\circ$, against 7.3\,cm\,/\,$0.7^\circ$ for a single manual calibration, its larger errors are mostly in rotation, and starting from a random pose in the search box changes the result little. In simulation its median error is 5.3\,cm\,/\,$0.19^\circ$; what remains there is mainly a bias of the cost, away from the camera and downwards, and the real estimates are offset away from the camera too. Future work includes a cost whose minimum lies at the true extrinsic, reference-free confidence estimation, and deployment on pole-mounted infrastructure.

The code and configurations are available at \url{https://github.com/zsungi/fgscalib}, and the dataset, with the foreground masks, point sets and fields of both evaluations and the estimates of the reported runs, at \url{https://doi.org/10.5281/zenodo.24448317}; camera recordings are anonymised. Further experiments and details are in the supplementary material.

\section*{Acknowledgment}
The authors used Claude (Anthropic) to help draft and edit the text of all sections and to help write the released code. The authors checked all content and are responsible for it.

\bibliographystyle{IEEEtran}

\bibliography{bibliography}

\begin{thebibliography}{10}
\providecommand{\url}[1]{#1}
\csname url@samestyle\endcsname
\providecommand{\newblock}{\relax}
\providecommand{\bibinfo}[2]{#2}
\providecommand{\BIBentrySTDinterwordspacing}{\spaceskip=0pt\relax}
\providecommand{\BIBentryALTinterwordstretchfactor}{4}
\providecommand{\BIBentryALTinterwordspacing}{\spaceskip=\fontdimen2\font plus
\BIBentryALTinterwordstretchfactor\fontdimen3\font minus
  \fontdimen4\font\relax}
\providecommand{\BIBforeignlanguage}[2]{{%
\expandafter\ifx\csname l@#1\endcsname\relax
\typeout{** WARNING: IEEEtran.bst: No hyphenation pattern has been}%
\typeout{** loaded for the language `#1'. Using the pattern for}%
\typeout{** the default language instead.}%
\else
\language=\csname l@#1\endcsname
\fi
#2}}
\providecommand{\BIBdecl}{\relax}
\BIBdecl

\bibitem{zhang2026roadside_review}
J.~Zhang, C.~Ge, W.~Xiao, M.~Tang, J.~Mills, B.~Coifman, and N.~Chen,
  ``Roadside lidar-based scene understanding toward intelligent traffic
  perception: A comprehensive review,'' \emph{ISPRS Journal of Photogrammetry
  and Remote Sensing}, vol. 233, pp. 69--88, 2026.

\bibitem{snapcat3d}
D.~F. Glas, D.~Br{\v{s}}{\v{c}}i{\'c}, T.~Miyashita, and N.~Hagita,
  ``{SNAPCAT-3D}: Calibrating networks of {3D} range sensors for pedestrian
  tracking,'' in \emph{2015 IEEE International Conference on Robotics and
  Automation (ICRA)}, 2015, pp. 712--719.

\bibitem{ISPRS}
L.~Kovács, M.~Kégl, and C.~Benedek, ``Real-time foreground segmentation for
  surveillance applications in {NRCS} {LiDAR} sequences,'' \emph{Int. Archives
  of the Photogrammetry, Remote Sensing and Spatial Information Sciences}, vol.
  XLIII-B1-2022, pp. 45--51, May 2022.

\bibitem{calibration_review}
Z.~Liu, Z.~Chen, X.~Wei, W.~Chen, and Y.~Wang, ``External extrinsic calibration
  of multi-modal imaging sensors: A review,'' \emph{IEEE Access}, vol.~11, pp.
  110\,417--110\,441, 2023.

\bibitem{tian2024rlcformer}
R.~Tian, X.~Bao, Y.~Chen, F.~Liu, Y.~Zhen, and Y.~Li, ``{RLCFormer}: Automatic
  roadside {LiDAR}-camera calibration framework with transformer,''
  \emph{Heliyon}, vol.~10, no.~20, p. e38506, 2024.

\bibitem{OLIVEIRA2022ATOM}
M.~de~Oliveira \emph{et~al.}, ``{ATOM}: A general calibration framework for
  multi-modal, multi-sensor systems,'' \emph{Expert Systems with Applications},
  vol. 207, p. 118000, 2022.

\bibitem{Pusztai_boxcalib_2017}
Z.~Pusztai and L.~Hajder, ``Accurate calibration of lidar-camera systems using
  ordinary boxes,'' in \emph{2017 IEEE International Conference on Computer
  Vision Workshops (ICCVW)}, 2017, pp. 394--402.

\bibitem{tekla_calibsphere_2020}
T.~Tóth, Z.~Pusztai, and L.~Hajder, ``Automatic lidar-camera calibration of
  extrinsic parameters using a spherical target,'' in \emph{2020 IEEE
  International Conference on Robotics and Automation (ICRA)}, 2020, pp.
  8580--8586.

\bibitem{semalign}
Z.~Liu, H.~Tang, S.~Zhu, and S.~Han, ``{SemAlign}: Annotation-free
  camera-{LiDAR} calibration with semantic alignment loss,'' in \emph{2021
  IEEE/RSJ International Conference on Intelligent Robots and Systems (IROS)},
  2021, pp. 8845--8851.

\bibitem{koide2023general}
K.~Koide, S.~Oishi, M.~Yokozuka, and A.~Banno, ``General, single-shot,
  target-less, and automatic lidar-camera extrinsic calibration toolbox,'' in
  \emph{2023 IEEE International Conference on Robotics and Automation (ICRA)},
  2023, pp. 11\,301--11\,307.

\bibitem{hku_calib}
C.~Yuan, X.~Liu, X.~Hong, and F.~Zhang, ``Pixel-level extrinsic self
  calibration of high resolution lidar and camera in targetless environments,''
  \emph{IEEE Robotics and Automation Letters}, vol.~6, no.~4, pp. 7517--7524,
  2021.

\bibitem{liu2022hku_adaptive_voxelization}
X.~Liu, C.~Yuan, and F.~Zhang, ``Targetless extrinsic calibration of multiple
  small fov lidars and cameras using adaptive voxelization,'' \emph{IEEE
  Transactions on Instrumentation and Measurement}, vol.~71, pp. 1--12, 2022.

\bibitem{sst-calib}
A.~Kodaira, Y.~Zhou, P.~Zang, W.~Zhan, and M.~Tomizuka, ``{SST-Calib}:
  Simultaneous spatial-temporal parameter calibration between {LiDAR} and
  camera,'' in \emph{2022 IEEE 25th International Conference on Intelligent
  Transportation Systems (ITSC)}, 2022, pp. 2896--2902.

\bibitem{autos_n1}
C.~Sun, Z.~Wei, W.~Huang, Q.~Liu, and B.~Wang, ``Automatic targetless
  calibration for lidar and camera based on instance segmentation,'' \emph{IEEE
  Robotics and Automation Letters}, vol.~8, no.~2, pp. 981--988, 2023.

\bibitem{nagyB_2020_calib}
B.~Nagy and C.~Benedek, ``On-the-fly camera and lidar calibration,''
  \emph{Remote Sensing}, vol.~12, no.~7, 2020.

\bibitem{motion_based_calib}
Z.~Taylor and J.~Nieto, ``Motion-based calibration of multimodal sensor
  arrays,'' in \emph{2015 IEEE International Conference on Robotics and
  Automation (ICRA)}, May 2015, pp. 4843--4850.

\bibitem{sfm_balazs_nagy}
B.~Nagy, L.~Kovács, and C.~Benedek, ``{SfM} and semantic information based
  online targetless camera-{LiDAR} self-calibration,'' in \emph{2019 IEEE
  International Conference on Image Processing (ICIP)}, 2019, pp. 1317--1321.

\bibitem{nagy2019online_e2e_calib}
B.~Nagy, L.~Kovács, and C.~Benedek, ``Online targetless end-to-end
  camera-lidar self-calibration,'' in \emph{2019 16th International Conference
  on Machine Vision Applications (MVA)}, 2019, pp. 1--6.

\bibitem{regnet}
N.~Schneider, F.~Piewak, C.~Stiller, and U.~Franke, ``{RegNet}: Multimodal
  sensor registration using deep neural networks,'' in \emph{2017 IEEE
  Intelligent Vehicles Symposium (IV)}, 2017, pp. 1803--1810.

\bibitem{lccnet}
X.~Lv, B.~Wang, Z.~Dou, D.~Ye, and S.~Wang, ``{LCCNet}: {LiDAR} and camera
  self-calibration using cost volume network,'' in \emph{2021 IEEE/CVF
  Conference on Computer Vision and Pattern Recognition Workshops (CVPRW)},
  2021, pp. 2888--2895.

\bibitem{zhang2025lcc_review}
H.~Zhang, S.~Li, X.~Zhu, H.~Chen, and W.~Yao, ``3-{D} {LiDAR} and monocular
  camera calibration: A review,'' \emph{IEEE Sensors Journal}, vol.~25, no.~7,
  pp. 10\,530--10\,555, 2025.

\bibitem{jang2025trafficcalib}
B.-J. Jang, T.-L. Kim, and T.-H. Park, ``Camera--{LiDAR} wide range calibration
  in traffic surveillance systems,'' \emph{Sensors}, vol.~25, no.~3, p. 974,
  2025.

\bibitem{sarlin2020superglue}
P.-E. Sarlin, D.~DeTone, T.~Malisiewicz, and A.~Rabinovich, ``{SuperGlue}:
  Learning feature matching with graph neural networks,'' in \emph{Proc.
  IEEE/CVF Conf. Comput. Vis. Pattern Recognit. (CVPR)}, 2020, pp. 4938--4947.

\bibitem{autos_regi}
Y.~Zhu, C.~Li, and Y.~Zhang, ``Online camera-lidar calibration with sensor
  semantic information,'' in \emph{2020 IEEE International Conference on
  Robotics and Automation (ICRA)}, 2020, pp. 4970--4976.

\bibitem{calibanything}
Z.~Luo, G.~Yan, and Y.~Li, ``Zero-training {LiDAR}-camera extrinsic calibration
  method using {Segment Anything Model},'' in \emph{2024 IEEE International
  Conference on Robotics and Automation (ICRA)}, 2024, pp. 14\,472--14\,478.

\bibitem{pairs2026}
H.-Y. Jeong, W.-H. Son, D.-W. Shin, K.~Cho, M.~Chee, and T.~Oh, ``Targetless
  {LiDAR}--camera extrinsic calibration via class-agnostic boundary mask
  alignment and {SPSA}-based optimization,'' \emph{Sensors}, vol.~26, no.~5, p.
  1501, 2026.

\bibitem{calibnet}
G.~Iyer, R.~K. Ram, J.~K. Murthy, and K.~M. Krishna, ``{CalibNet}:
  Geometrically supervised extrinsic calibration using {3D} spatial transformer
  networks,'' in \emph{2018 {IEEE}/{RSJ} International Conference on
  Intelligent Robots and Systems ({IROS})}, 2018, pp. 1110--1117.

\bibitem{cmrnext}
D.~Cattaneo and A.~Valada, ``{CMRNext}: Camera to {LiDAR} matching in the wild
  for localization and extrinsic calibration,'' \emph{IEEE Transactions on
  Robotics}, vol.~41, pp. 1995--2013, 2025.

\bibitem{raft}
Z.~Teed and J.~Deng, ``{RAFT}: Recurrent all-pairs field transforms for optical
  flow,'' in \emph{European Conference on Computer Vision (ECCV)}.\hskip 1em
  plus 0.5em minus 0.4em\relax Springer, 2020, pp. 402--419.

\bibitem{huang2025whatmatters}
S.~Huang, C.~Lin, and Y.~Zhao, ``What really matters for learning-based
  {LiDAR}-camera calibration,'' in \emph{Proceedings of the ACM International
  Conference on Multimedia in Asia (MMAsia)}, 2025.

\bibitem{3dgs_calib}
Q.~Herau, M.~Bennehar, A.~Moreau, N.~Piasco, L.~Roldao, D.~Tsishkou,
  C.~Migniot, P.~Vasseur, and C.~Demonceaux, ``{3DGS-Calib}: {3D} {Gaussian}
  splatting for multimodal spatiotemporal calibration,'' in \emph{2024 IEEE/RSJ
  International Conference on Intelligent Robots and Systems (IROS)}, 2024, pp.
  8315--8321.

\bibitem{pervsic_trajectory_based}
J.~Per{\v{s}}i{\'c}, L.~Petrovi{\'c}, I.~Markovi{\'c}, and I.~Petrovi{\'c},
  ``Online multi-sensor calibration based on moving object tracking,''
  \emph{Advanced Robotics}, vol.~35, no. 3-4, pp. 130--140, 2021.

\bibitem{Tsai2021OptimisingTS}
D.~Tsai, S.~Worrall, M.~Shan, A.~Lohr, and E.~Nebot, ``Optimising the selection
  of samples for robust lidar camera calibration,'' in \emph{2021 IEEE
  International Intelligent Transportation Systems Conference (ITSC)}, 2021,
  pp. 2631--2638.

\bibitem{zivkovic2004improved_MOG2}
Z.~Zivkovic, ``Improved adaptive {Gaussian} mixture model for background
  subtraction,'' in \emph{Proceedings of the 17th International Conference on
  Pattern Recognition (ICPR)}, vol.~2, 2004, pp. 28--31.

\bibitem{stauffer_grimson}
C.~Stauffer and W.~Grimson, ``Adaptive background mixture models for real-time
  tracking,'' in \emph{Proc. IEEE Conf. Computer Vision and Pattern Recognition
  (CVPR)}, vol.~2, 1999, pp. 246--252.

\bibitem{borgefors1988hierarchical}
G.~Borgefors, ``Hierarchical chamfer matching: A parametric edge matching
  algorithm,'' \emph{IEEE Transactions on Pattern Analysis and Machine
  Intelligence}, vol.~10, no.~6, pp. 849--865, 1988.

\bibitem{levinson2013automatic}
J.~Levinson and S.~Thrun, ``Automatic online calibration of cameras and
  lasers,'' in \emph{Robotics: Science and Systems (RSS)}, 2013.

\bibitem{borgefors1986distance}
G.~Borgefors, ``Distance transformations in digital images,'' \emph{Computer
  Vision, Graphics, and Image Processing}, vol.~34, no.~3, pp. 344--371, 1986.

\bibitem{kirkpatrick1983sa}
S.~Kirkpatrick, C.~D. Gelatt, and M.~P. Vecchi, ``Optimization by simulated
  annealing,'' \emph{Science}, vol. 220, no. 4598, pp. 671--680, 1983.

\bibitem{geiger2013vision}
A.~Geiger, P.~Lenz, C.~Stiller, and R.~Urtasun, ``Vision meets robotics: The
  {KITTI} dataset,'' \emph{The International Journal of Robotics Research},
  vol.~32, no.~11, pp. 1231--1237, 2013.

\bibitem{dairv2x}
H.~Yu, Y.~Luo, M.~Shu, Y.~Huo, Z.~Yang, Y.~Shi, Z.~Guo, H.~Li, X.~Hu, J.~Yuan,
  and Z.~Nie, ``{DAIR-V2X}: A large-scale dataset for vehicle-infrastructure
  cooperative {3D} object detection,'' in \emph{Proceedings of the IEEE/CVF
  Conference on Computer Vision and Pattern Recognition (CVPR)}, 2022, pp.
  21\,361--21\,370.

\bibitem{dosovitskiy2017carla}
A.~Dosovitskiy, G.~Ros, F.~Codevilla, A.~Lopez, and V.~Koltun, ``{CARLA}: An
  open urban driving simulator,'' in \emph{Conference on Robot Learning
  (CoRL)}, 2017, pp. 1--16.

\bibitem{kato2018autoware}
S.~Kato \emph{et~al.}, ``Autoware on board: Enabling autonomous vehicles with
  embedded systems,'' in \emph{ACM/IEEE Int.\ Conf.\ on Cyber-Physical Systems
  (ICCPS)}, 2018, pp. 287--296.

\bibitem{vim2012}
{Joint Committee for Guides in Metrology}, ``International vocabulary of
  metrology -- {B}asic and general concepts and associated terms ({VIM}),''
  BIPM, Tech. Rep. JCGM 200:2012, 2012.

\bibitem{jcgm100gum}
{Joint Committee for Guides in Metrology}, ``Evaluation of measurement data --
  {G}uide to the expression of uncertainty in measurement,'' BIPM, Tech. Rep.
  JCGM 100:2008, 2008.

\end{thebibliography}

\clearpage\thispagestyle{empty}\AddToShipoutPictureBG*{\AtPageLowerLeft{\includegraphics[page=1]{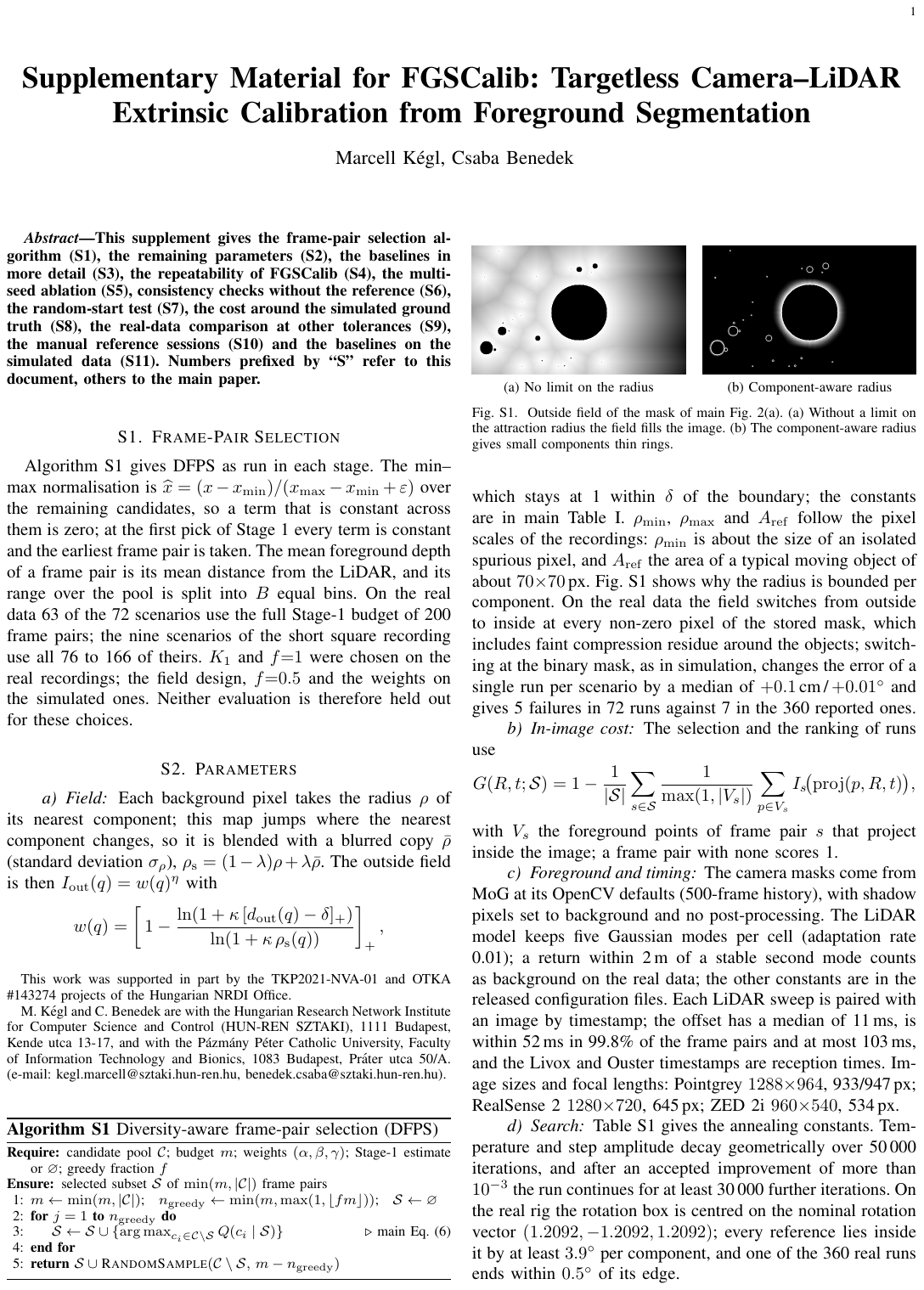}}}\null
\clearpage\thispagestyle{empty}\AddToShipoutPictureBG*{\AtPageLowerLeft{\includegraphics[page=2]{supplement.pdf}}}\null
\clearpage\thispagestyle{empty}\AddToShipoutPictureBG*{\AtPageLowerLeft{\includegraphics[page=3]{supplement.pdf}}}\null
\clearpage\thispagestyle{empty}\AddToShipoutPictureBG*{\AtPageLowerLeft{\includegraphics[page=4]{supplement.pdf}}}\null

\end{document}